\documentclass[10pt,twocolumn]{IEEEtrans}

\usepackage{graphicx}
\usepackage{amssymb}
\usepackage{rotating}

\newtheorem{definition}{Definition}
\newtheorem{example}{Example}

\begin{document}

\title{On selection of centroids of fuzzy clusters for color classification\thanks{This work was supported by the Korea Science and
Engineering Foundation (KOSEF)  through the Advanced Information
Technology Research Center}}
\author{Dae-Won Kim$^{1}$ and Kwang H. Lee$^{1,2}$
\\ Department of Electrical Engineering and Computer Science$^{1}$
\\ Department of BioSystems$^{2}$
\\ KAIST, Kusung-dong, Yusung-gu, 305-701, Daejon, Korea}

\maketitle

\pagenumbering{arabic}

\begin{abstract}
A novel initialization method in the fuzzy c-means (FCM) algorithm
is proposed for the color clustering problem. Given a set of color
points, the proposed initialization extracts dominant colors that
are the most vivid and distinguishable colors. Color points
closest to the dominant colors are selected as initial centroids
in the FCM. To obtain the dominant colors and their closest color
points, we introduce reference colors and define a fuzzy
membership model between a color point and a reference color.
\end{abstract}

\begin{keywords}
Fuzzy clustering, Color clustering, Centroid initialization, Fuzzy
c-means, Color membership
\end{keywords}

\section{Introduction}
\label{sec:introduction}

The objective of color clustering is to partition a color set into
$c$ homogeneous color clusters. This kind of color clustering task
can be widely used in a variety of applications such as color
image segmentation. Color clustering is inherently ambiguous task
and color boundaries are often blurred and truly
fuzzy~\cite{Bez99}. For example, consider the task of segmenting
an color image into color objects. The boundaries between objects
are blurred and distorted due to the imaging process. Furthermore,
object definitions are not always crisp, and knowledge about the
objects in the scene can be described on vague terms. Fuzzy set
theory and fuzzy logic are ideally suited for dealing with such
uncertainty. The fuzzy clustering approach is to preserve the
uncertainty inherent in the problem as long as possible until
actual decisions have to be made, which makes it possible  to be
less prone to local optimal than crisp clustering
algorithms~\cite{Bez99}\cite{Jai99}\cite{Sha00}\cite{Lee95}\cite{Lee97}\cite{Qzd02}\cite{Jai98}.

The most widely used fuzzy clustering algorithm is the fuzzy
c-means (FCM) algorithm proposed by Bezdek~\cite{Bez81}. This
algorithm classifies a collection of pattern data set $X$ into $c$
homogeneous groups represented as fuzzy sets
$\tilde{F}_1,\tilde{F}_2,...,\tilde{F}_c$. The objective is to
obtain the fuzzy $c$-partition $\tilde{F}=\{
\tilde{F}_1,\tilde{F}_2,..,\tilde{F}_c\}$ for both an unlabeled
data set $X=\{ x_1, ...,x_n\}$ and the number of clusters $c$ by
minimizing the function $J_m$

\begin{equation}
minimize~
J_m(U,V:X)=\sum_{i=1}^{c}\sum_{j=1}^{n}(\mu_{ij})^{m}\|x_j-v_i\|^2
\end{equation}

where $\mu_{ij}$ is the membership degree of data $x_j$ to a fuzzy
cluster set $\tilde{F}_i$, and also, is an element of a $(c \times
n)$ pattern matrix $U=[\mu_{ij}]$. An $i$-th row $U_i$ of $U$
corresponds to a fuzzy cluster set $\tilde{F}_i$.
$V=(v_1,v_2,..,v_c)$ is a vector of centroids of the fuzzy cluster
$(\tilde{F}_1,\tilde{F}_2,...,\tilde{F}_c)$. Thus, a fuzzy
partition can be denoted by the pair $(U,V)$. The $\|x_j-v_i\|$ is
an Euclidean norm between $x_j$ and $v_i$. The parameter $m$
controls the fuzziness of membership of each datum. The goal is to
iteratively improve a sequence of sets of fuzzy clusters
$\tilde{F}(1),\tilde{F}(2),...\tilde{F}(t)$($t$ means the
iteration step) until no further improvement in $J_m(U,V:X)$ is
possible.

However, as Gath and Geva pointed out~\cite{Gat89}\cite{Iye00},
there are three major difficulties in fuzzy clustering: (1)
finding the number of clusters to be clustered. Most algorithms
require the user to specify the number of clusters. (2) the choice
of the initial centroids of clusters. Most algorithms choose an
arbitrary random selection because they converge for sure after
the iterative process. (3) a method to tackle data that has a
large variability in cluster shapes, variation in cluster
densities and different number of points in different clusters, is
needed. This paper concerns with the second difficulty, i.e.,
initialization of clustering. The initialization is very
important. The algorithm requires the starting points, i.e., the
initial centroids of clusters, to run the iteration; however, it
is not always possible to know the exact initial centroids in
advance. Different fuzzy clusters might be obtained according to
the selected starting points. There is no general agreement on a
good initialization scheme. The three most popular techniques are:
(1) using $c$ data points randomly selected in the set (2) using
the first $c$ distinct data points in the set; (3) using $c$ data
points uniformly distributed across the data set
~\cite{Bez99}\cite{The97}\cite{Lee98}. To tackle this, we attempt
to determine dominant colors for a given color set, which are
obtained by computing the similarities between color points in the
set and well-known reference colors. The similarity is given a
value of the degree of membership to reference colors. The
dominant colors are used as clues in the selection of the initial
centroids.

\section{Membership function of color}
\label{sec:membership-function-of-color}

There is no general agreement on a good initialization scheme. We
present a novel approach to select the initial centroids of
clusters when the FCM algorithm is applied to the color clustering
problem. It is intuitive and straightforward. The key factor is to
search for colors that are dominant and distinguishable for a
given color data set. We assume that the dominant and vivid colors
would help to separate the color space and not belong to the same
cluster. Thus, the dominant colors can be used to guess the
initial starting points. In addition, in this paper, we selected
CIELAB color space for color clustering. Perceptually uniform
spaces such as CIELAB color space have been used in some instances
for image processing, with the main justifications for both their
use being their intuitive appeal to humans and their superior
performance~\cite{Pas01}\cite{Gon98}\cite{Cha96}\cite{Li00}\cite{Sha98}.

Given a color point $x \in X$, a computational model of membership
function to other colors is required to determine dominant colors.
Let us consider the following two factors for this.

\begin{itemize}
\item a set of reference colors
\item a (dis)similarity measure between a color point and
reference colors
\end{itemize}

First, we suppose that a set of reference colors contains all
major colors that can be observed in natural scenes. This provides
a standard with which to compare and measure similarities between
colors. Typically, 10-20 colors are needed in the images of
natural scenes~\cite{Den99}. For this purpose, we adopted 14
reference colors presented in ColorChecker chart that is produced
in the Munsell Color Lab at
GretagMacbeth~\cite{Lon94}\cite{Luk96}\cite{Mun03}. The
ColorChecker chart includes scientifically designed various
colors, which represent natural colors under any illumination,
determine the true color distribution of color space, and
reproduce all colors perfectly. This has been used as a reference
guide for color selection for a variety of
applications~\cite{Luk96}. Table 1 lists the 14 reference colors
and their CIELAB values selected in this paper.

\begin{table}[t]
\label{table:reference-color}
\begin{center}
\caption{Reference colors and their CIELAB values}
\begin{tabular}{lccc}
\hline Reference color & $L^*$ & $a^*$ & $b^*$ \\
\hline
$R_1$ Red               & 41.34 &  49.31 &  24.65\\
$R_2$ Green             & 55.03 &  -40.14 & 32.29\\
$R_3$ Blue              & 30.35 &  26.44 &  -49.67 \\
$R_4$ Yellow            & 80.70 & -3.66 &  77.55\\
$R_5$ Magenta           & 51.14 &  48.16 &  -15.29\\
$R_6$ Cyan              & 51.15 & -19.72 & -23.38\\
$R_7$ Dark skin         & 38.02 & 11.80 & 13.66\\
$R_8$ Orange            & 61.13 & 28.11 & 56.13 \\
$R_9$ Purple            & 31.10 & 24.36 & -22.11\\
$R_{10}$ Greenish yellow   & 71.90 & -28.11 & 56.96\\
$R_{11}$ Bluish green     & 71.00 & -30.63 & 1.53\\
$R_{12}$ Light skin     & 65.67 & 13.68 & 16.89\\
$R_{13}$ Black          & 0.00 & 0.00 & 0.00\\
$R_{14}$ White          & 95.82 & -0.17 &  0.47\\ \hline
\end{tabular}
\end{center}
\end{table}

Second, the similarity measure is an indicator of how similar a
color point is to reference colors. We see that color groups do
not have crisp boundaries and similarity measures between color
data are not clear. Thus, a soft or fuzzy similarity measure is
more appropriate than a hard or crisp similarity. Suppose that a
color point, denoted by $x \in X$, is a point on CIELAB color
space, which is represented by ($x_L, x_a, x_b$) tuple value. A
reference color, denoted by $R_i \in R$, is a point represented by
($r^i_L, r^i_a, r^i_b$) tuple value on CIELAB space as shown in
Table 1. For a given color point $x$, $\mu_{R_i}: x \rightarrow
[0,1]$ quantifies the degree of membership $\mu_{R_i}(x)$ of $x$
to the reference color $R_i$, which expresses how similar the
color point $x$ is to the reference color $R_i$.

To address the definition of the membership computation model, we
define a distance measure between a color point and a reference
color.

\vspace{3mm}
\begin{definition}
Let $x = (x_L,x_a,x_b)$ and $R_i = (r^i_L,r^i_a,r^i_b)$ be a color
point and a reference color on CIELAB color space, respectively.
Then the distance between $x$ and $R_i$, denoted by
$\delta(x,R_i)$, is defined as
\begin{equation}
\delta(x,R_i) = \sqrt{(x_L-r^i_L)^2+(x_a-r^i_a)^2+(x_b-r^i_b)^2}
\end{equation}
\end{definition}
\vspace{3mm}

The above definition is obvious from the CIELAB color difference
formula~\cite{Wys00}. With the distance measure, we formulate the
membership computation model between a color point and a reference
color.

\vspace{3mm}
\begin{definition}~\label{def:membership-function} Let $x$ and $R_i$ be a color point and a reference color on CIELAB
color space, respectively. Let $R = \{ R_1, R_2, ..., R_k \}$ be a
universe of discourse of reference colors. Then the membership
function of $x$ to a reference color $R_i$ is defined as
\begin{equation}
\mu_{R_i}(x)= \left \{
            \begin{array}{ll}
                1.0 & \mbox{if~ $\delta(x, R_i) = 0$} \\
                0.0 & \mbox{if~ $\delta(x, R_j) = 0$} \\
                \displaystyle \Bigg(\sum_{j=1}^{k}\bigg(\frac{\delta(x, R_i)}{\delta(x,R_j)}\bigg)^{\lambda}\Bigg)^{-1} &
                \mbox{otherwise}
            \end{array}
            \right.
\end{equation}
\end{definition}
\vspace{3mm}

where  the parameter $\lambda$ controls the fuzziness of
membership of $x$ to $R_i$. If a color point $x$ is equal to a
reference color $R_i$, it has a membership degree of 1.0.
Conversely, if $x$ is equal to other reference color $R_j$, then
it means $x$ has no relation to $R_i$, thus the membership degree
is given a value of 0.0. Except for the above two cases, the color
point $x$ is located between reference colors. We compute relative
distances from reference colors and determine the membership
value. The sum of membership values to the whole reference colors
must be equal to 1.0. We see that the following properties
obviously hold from Definition~\ref{def:membership-function}.

\vspace{3mm}
\begin{itemize}
\item $0 \leq \mu_{R_i}(x) \leq 1$.

\item $\sum_{i=1}^{k}\mu_{R_i}(x)=1.0$

\item If $\delta(x, R_i) < \delta(x, R_j)$, then $\mu_{R_i}(x) \geq
\mu_{R_j}(x)$.
\end{itemize}

\vspace{3mm}
\begin{example}
Let $x$ be a color point, and let us consider three reference
colors $R_{red}, R_{green}$, and $R_{blue}$. For a color element
$x$ and $\lambda=1.0$, consider
$$
\delta(x, R_{red}) = 20, \quad \delta(x, R_{green}) = 30, \quad
\delta(x, R_{blue}) = 10
$$
Then the degree of membership $x$ to $R_{red}$ is calculated as:
\begin{eqnarray*}
\lefteqn{\mu_{R_{red}}(x)} \\
& & =\Bigg(
\bigg(\frac{\delta(x,R_{red})}{\delta(x,R_{red})}\bigg) +
\bigg(\frac{\delta(x,R_{red})}{\delta(x,R_{green})}\bigg) +
\bigg(\frac{\delta(x,R_{red})}{\delta(x,R_{blue})}\bigg)\Bigg)^{-1} \\
& & =(20/20 + 20/30 + 20/10)^{-1} \\ & & =0.27
\end{eqnarray*}
We see that the color point $x$ is similar to the reference color
$R_{red}$ with a membership 0.27. In a similar way, we calculate
the other two membership degrees $\mu_{R_{green}}(x) = 0.18$ and
$\mu_{R_{blue}}(x) = 0.55$. Thus the reference color `Blue' is the
closest to $x$.
\end{example}

\section{The Proposed initialization of clustering}
\label{sec:membership-function-of-color}

The initialization of clustering is to find good starting points
for the initial centroids of clusters. As mentioned earlier, we
guess the starting points from the dominant colors. The dominant
colors are a subset of reference colors that record the highest
degrees of membership for all $x \in X$. The number of dominant
colors are equal to the number of groups to be clustered.

In order to determine dominant colors, we compute degrees of
membership between color points $x_j \in X$ and reference colors
$R_i \in R$. A reference color $R_i$ has two additional attributes
denoted by $\mu_{i}$ and $p_{i}$. The reference color thus defined
as
\begin{equation}\label{eq:reference-color}
R_{i} = \{ \{ r^i_L, r^i_a, r^i_b \} , ~\mu_i , ~p_i  \}
\end{equation}
where
\begin{eqnarray*}
\mu_i & = & max \{ ~\mu_{R_i}(x_j) ~\} \quad \mbox{for all $x_j \in X$} \\
p_i & = & \{ x_j ~|~ \mu_{R_i}(x_j) \geq \mu_{R_i}(x_k) \}\quad
\mbox{for all $x_j, x_k \in X$}
\end{eqnarray*}

In Equation~\ref{eq:reference-color}, $\{ r^i_L, r^i_a, r^i_b \}$
is a CIELAB value, $\mu_{i}$ indicates the highest degree of
membership obtained by computing $\mu_{R_i}(x_j)$ for all $x_j \in
X$, and $p_i$ indicates the closest color point $x_j$ to $R_i$.
For a color point $x_j \in X$, we compute the color membership
degree to the 14 reference colors, and update $\mu_i$ and $p_i$.
When the computation is completed, the reference colors are sorted
by $\mu_i$ in descending order. The sorted list of reference
colors $R^s$ is represented by
\begin{equation}
R^s = ( R^s_1,R^s_2,...,R^s_k )
\end{equation}
where the reference color $R^s_1$ has the highest value of $\mu_i$
and $R^s_k$ has the lowest one. Now we can define the dominant
colors below.

\vspace{3mm}
\begin{definition}\label{def:dominant-color} Let $R$ and $R^s$ be a set of reference colors and a sorted list of reference colors, respectively. The number of groups to be
clustered is denoted by $c$. Then a set of dominant colors $D = \{
D_1, D_2, ..., D_c \}$ is defined as
\begin{equation}
D = \{ D_i ~|~ D_i = R^s_i , ~ 1 \leq i \leq c  \}
\end{equation}
\end{definition}
\vspace{3mm}

We see that a set of dominant colors consists of the first $c$
reference colors in the sorted list, which represent the most
distinguishable and vivid colors in a given color set $X$. When
dominant colors are determined, starting points for the initial
centroids of clusters are obtained as below

\vspace{3mm}
\begin{definition}\label{def:initial-centroid}
Let $D=\{D1,...,D_c\}$ be a set of dominant colors. Let c be the
number of groups to be clustered. Then the initial centroids
$V_0=\{ v_1, v_2,...,v_c\}$ is defined as
\begin{equation}
V_0 = \{ v_i ~|~ v_i = p_i, ~ p_i \in D_i \}
\end{equation}
\end{definition}
\vspace{3mm}

where $v_{i}$ is given the color point $p_i$ that is closest to
the dominant color $D_i$.

We now describe the procedural steps for the FCM algorithm using
the proposed initialization method.

\begin{description}

\vspace{2mm}
\item[Step 1.] Given a number of clusters $c$, choose the
parameters of the FCM and the initialization: a termination
criterion $\epsilon$, weighting exponents $m$ and $\lambda$.

\vspace{2mm}
\item[Step 2.] Initialize color points $x_j \in X$ for $j=1,2,..,n$ and reference colors $R_i \in R$ for
$i=1,2,...,k=14$.

\vspace{2mm}
\item[Step 3.] For each $x_j \in X$, compute degrees of membership to
all reference colors $R_i$'s,
\begin{equation}
\mu_{R_i}(x_j)= \left \{
            \begin{array}{ll}
                1.0 & \mbox{if~ $\delta(x_j, R_i) = 0$} \\
                0.0 & \mbox{if~ $\delta(x_j, R_j) = 0$}\\
                \displaystyle \Bigg(\sum_{j=1}^{k}\bigg(\frac{\delta(x_j, R_i)}{\delta(x_j,R_j)}\bigg)^{\lambda}\Bigg)^{-1} &
                \mbox{otherwise}
            \end{array}
            \right.
\end{equation}
and update $\mu_i$ and $p_i$ of $R_i$
\begin{eqnarray*}
\mu_i & = & max \{ ~\mu_{R_i}(x_j) ~\} \quad \mbox{for all $x_j \in X$} \\
p_i & = & \{ x_j ~|~ \mu_{R_i}(x_j) \geq \mu_{R_i}(x_k) \}\quad
\mbox{for all $x_j, x_k \in X$}
\end{eqnarray*}

\vspace{2mm}
\item[Step 4.] Sort the reference colors $R_i \in R$ in descending order
of $\mu_i$, which yields a sorted list $R^s$.

\vspace{2mm}
\item[Step 5.] Determine dominant colors $D_1,D_2,...,D_c$ from $R^s$ and
assign $p_i \in D_i$ to the initial centroids
$v_1(0),v_2(0),...,v_c(0)(t=0)$.

\vspace{2mm}
\item[Step 6.] Compute the fuzzy cluster $\tilde{F}_i(t)$ for $i=1,2,...,c$. For each
$x_j$:
\begin{equation}
\mu_{\tilde{F}_i}(x_j)(t) =
\Bigg(\sum_{k=1}^{c}\bigg(\frac{\|x_j-v_i\|^2}{\|x_j-v_k\|^2}\bigg)^{\frac{1}{m-1}}\Bigg)^{-1}
\end{equation}

\vspace{2mm}
\item[Step 7.] Update the fuzzy cluster centroid $v_i(t+1)$ for $i=1,2,...,c$ using
\begin{equation}
v_i(t+1) =
\frac{\sum_{j=1}^{n}(\mu_{\tilde{F}_i}(x_j))^{m}x_j}{\sum_{j=1}^{n}(\mu_{\tilde{F}_i}(x_j))^m}
\end{equation}

\vspace{2mm}
\item[Step 8.] If $|\tilde{F}_i(t+1)-\tilde{F}_i(t)| < \epsilon$ for all $\tilde{F}_i\in\tilde{F}$, then stop. We found no further improvement
in $J(\tilde{F})$, and believed that the algorithm has reached at
convergence; otherwise, $t \leftarrow t+1$ and go to step 6.

\end{description}

\vspace{3mm}
\begin{example}
As a simple illustration, we consider a color data set $X$
consisting of 10 color points in Table 2. The number of groups to
be clustered is given as three ($c=3$). According to the proposed
initialization, degrees of membership of each data to the 14
reference colors are calculated. Table 3 lists degrees of
membership of each data to the reference colors where each $\mu_i$
of $R_i$ is marked by bold.
\end{example}

\begin{table}[t]
\label{table:color-points}
\begin{center}
\caption{Sample color points and their CIELAB values}
\begin{tabular}{cccc}
\hline Color element & $L^*$ & $a^*$ & $b^*$ \\
\hline
$x_1$ & 32.65 & 55.94 & 28.89\\
$x_2$ & 41.37 & 61.45 & 34.49\\
$x_3$ & 8.98 & -1.02 & -2.73\\
$x_4$ & 87.63 & -20.77 & 68.00\\
$x_5$ & 41.01 & 45.03 & 20.65\\
$x_6$ & 80.70 & -5.76 & 70.55\\
$x_7$ & -8.04 & -1.00 & -30.34\\
$x_8$ & 85.71 & -15.87 & 67.65\\
$x_9$ & -8.26 & 0.00 & 0.05\\
$x_{10}$ & 5.00 & 2.27 & 1.52\\ \hline
\end{tabular}
\end{center}
\end{table}

\begin{table*}[t]
\label{table:refMV}
\begin{center}
\caption{Degrees of membership of $x_j$ to 14 reference colors}
\begin{tabular}{c|cccccccccc}
\hline $R_i$    & $x_1$         & $x_2$         & $x_3$         & $x_4$         & $x_5$         & $x_6$         & $x_7$         & $x_8$         & $x_9$ & $x_{10}$  \\
\hline
$R_1$     &0.71           &0.61           &0.02           &0.02           &\textbf{0.87}  &0.01           &0.04           &0.02           &0.01           &0.01 \\
$R_2$     &0.01           &0.01           &0.01           &\textbf{0.06}  &0.00           &0.01           &0.03           &0.05           &0.01           &0.01 \\
$R_3$     &0.01           &0.02           &0.02           &0.01           &0.01           &0.00           &\textbf{0.12}  &0.01           &0.01           &0.01 \\
$R_4$     &0.01           &0.02           &0.01           &0.35           &0.00           &\textbf{0.84}  &0.02           &0.46  &0.00           &0.00 \\
$R_5$     &0.04           &\textbf{0.05}  &0.02           &0.01           &0.02           &0.00           &0.05           &0.01           &0.01           &0.01\\
$R_6$     &0.01           &0.01           &0.03           &0.02           &0.00           &0.00           &\textbf{0.08}  &0.01           &0.01           &0.01 \\
$R_7$     &0.04           &0.05           &0.05           &0.02           &0.03           &0.01           &\textbf{0.07}  &0.02           &0.02           &0.02 \\
$R_8$     &0.04           &\textbf{0.08}  &0.01           &0.05           &0.02           &0.03           &0.02           &0.05           &0.01           &0.00\\
$R_9$     &0.03           &0.03           &0.04           &0.01           &0.01           &0.00           &0.14           &0.01           &\textbf{0.32}           &0.02 \\
$R_{10}$  &0.01           &0.02           &0.01           &\textbf{0.36}  &0.00           &0.06           &0.02           &0.28           &0.01           &0.00\\
$R_{11}$  &0.01           &0.01           &0.01           &0.03           &0.00           &0.01           &\textbf{0.04}  &0.03           &0.01           &0.01 \\
$R_{12}$  &0.03           &\textbf{0.05}  &0.02           &0.04           &0.02           &0.01           &0.04           &0.03           &0.01           &0.01\\
$R_{13}$  &0.02           &0.02           &0.75           &0.01           &0.01           &0.00           &0.31           &0.01           &0.86           &\textbf{0.90} \\
$R_{14}$  &0.01           &0.02           &0.01           &\textbf{0.03}  &0.01           &0.01           &0.03           &0.03           &0.01           &0.00\\
\hline
\end{tabular}
\end{center}
\end{table*}

Table 4 lists a summary on the reference colors and their
attributes obtained from Table 3. From the summary, we can obtain
a sorted list of reference colors $R^s = (Black, Red, Yellow, ...,
White )$ where Black has the highest value of $\mu_i$ and White
has the lowest value. Thus, the dominant colors are determined by
the first three colors, Black, Red, and Yellow in the sorted list.
We consider these three colors as the most distinguishable colors
in $X$, and choose the $p_i$'s of these colors as the starting
initial centroids: $x_{10}$ for Black, $x_5$ for Red, and $x_6$
for Yellow. Table 5 shows the three initial centroids $v_i$ for
$i=1,2,3$.

\begin{table}[t]
\label{table:output-set}
\begin{center}
\caption{Summary of reference colors $R$ and their attributes}
\begin{tabular}{lccccc}
\hline Reference color & $L^*$ & $a^*$ & $b^*$ & $\mu_i$ & $p_i$ \\
\hline
$R_1$  \textbf{Red} & 41.34 & 49.31 & 24.65    & \textbf{0.87} & $x_5$\\
$R_2$  Green   & 55.03 & -40.14 & 32.29   & 0.06 & $x_4$\\
$R_3$  Blue  & 30.35 & 26.44 & -49.67   & 0.12 & $x_7$\\
$R_4$  \textbf{Yellow} & 80.70 & -3.66 & 77.55    & \textbf{0.84} & $x_6$\\
$R_5$  Magenta  & 51.14 & 48.16 & -15.29   & 0.05 & $x_2$\\
$R_6$  Cyan  & 51.15 & -19.72 & -23.38  & 0.08 & $x_{7}$\\
$R_7$  Dark skin  & 38.02 & 11.80 & 13.66    & 0.07 & $x_{7}$\\
$R_8$  Orange  & 61.13 & 28.11 & 56.13    & 0.08 & $x_2$\\
$R_9$  Purple  & 31.10 & 24.36 & -22.11   & 0.32 & $x_9$\\
$R_{10}$ Greenish yellow & 71.90 & -28.11 & 56.96   & 0.36 & $x_{4}$\\
$R_{11}$ Bluish green & 71.00 & -30.63 & 1.53    & 0.04 & $x_{7}$\\
$R_{12}$ Light skin & 65.67 & 13.68 & 16.89    & 0.05 & $x_{2}$\\
$R_{13}$ \textbf{Black} & 0.00  & 0.00 & 0.00      & \textbf{0.90} & $x_{10}$\\
$R_{14}$ White & 95.82 & -0.17 & 0.47     & 0.03 & $x_{4}$\\
\hline
\end{tabular}
\end{center}
\end{table}

\begin{table}[t]
\label{table:initial-centroid}
\begin{center}
\caption{Initial centroids for three clusters}
\begin{tabular}{ccccc}\hline
Initial centroid & Color point & $L^*$ & $a^*$ & $b^*$\\\hline
$v_1$ & $x_{10}$ & 5.00 & 2.27 & 1.52 \\
$v_2$ & $x_5$ & 41.01 & 45.03 & 20.65 \\
$v_3$ & $x_6$ & 80.70 & -5.76 & 70.55 \\\hline
\end{tabular}
\end{center}
\end{table}

\section{Conclusion}
\label{sec:conclusion}

This paper discussed the initialization of the fuzzy c-means
clustering algorithm. As noted, different selections of the
initial centroids of clusters might lead to different local optima
or different partitions. We proposed to exploit dominant colors in
order to select the initial points. Dominant colors are expected
to be the most distinguishable colors, and belong to each separate
clusters. To accomplish this, reference colors were introduced and
a membership model for color was defined. Color points closest to
the dominant colors are finally selected as initial centroids.

\vspace{5mm}
\textbf{Dae-Won Kim} received the B.S. degree from
the Department of Computer Engineering, Kyungpook National
University, Daegu, Korea, in 1997, the M.S. degree from the
Department of Computer Science, KAIST, Daejon, in 1993,
respectively. He is currently pursuing the Ph.D. degree in
computer science at KAIST. His current interest fields are fuzzy
theory and applications, cluster analysis, and color image
management.

\vspace{5mm}
\textbf{Kwang H. Lee} received the B.S. degree from
the Department of Industrial Engineering, Seoul National
University, Seoul, Korea, in 1978, the M.S. degree from the
Department of Industrial Engineering, KAIST, Daejon, in 1980, the
D.E.A and Dr.Ing. degrees from the Department of Computer Science,
INSA de Lyon University, France, in 1982 and 1985, respectively,
and the Dr.Etat degree from the Department of Computer Science,
INSA de Lyon University, France, in 1988. He is currently a
professor at KAIST. His research interests include fuzzy systems,
artificial intelligence, and bioinformatics.

\end{document}